%% file: main.tex
\documentclass[10pt,twocolumn,letterpaper]{article}

\usepackage[pagenumbers]{cvpr} % To force page numbers, e.g. for an arXiv version
\usepackage{xcolor}
\usepackage{colortbl}
\definecolor{cvprblue}{rgb}{0.21,0.49,0.74}
\usepackage{multirow}
\usepackage{graphicx}

\makeatletter
\newcommand{\thickhline}{%
    \noalign {\ifnum 0=`}\fi \hrule height 1pt
    \futurelet \reserved@a \@xhline
}
\makeatother
\usepackage[pagebackref,breaklinks,colorlinks,allcolors=cvprblue]{hyperref}
\newcommand{\J}{$\mathcal{J}$\xspace}
\newcommand{\F}{$\mathcal{F}$\xspace}
\newcommand{\JF}{$\mathcal{J}\&\mathcal{F}$\xspace}

\newcommand{\JFnd}{$\mathcal{J}\&\dot{\mathcal{F}}_d$\xspace}
\newcommand{\JFnr}{$\mathcal{J}\&\dot{\mathcal{F}}_r$\xspace}
\newcommand{\Fn}{$\dot{\mathcal{F}}$\xspace}
\newcommand{\JFn}{$\mathcal{J}\&\dot{\mathcal{F}}$\xspace}

\def\paperID{*****} % *** Enter the Paper ID here
\def\confName{CVPR}
\def\confYear{2025}

\title{2nd Place Report of MOSEv2 Challenge 2025: Concept Guided Video Object Segmentation via SeC}

\author{
Zhixiong Zhang$^{1,2,3}$, Shuangrui Ding$^{4}$, Xiaoyi Dong$^{4}$, Yuhang Zang$^{3}$, Yuhang Cao$^{3}$, Jiaqi Wang$^{3}$\\
\small $^1$Shanghai Jiao Tong University \quad $^2$Shanghai Innovation Institute \quad $^3$Shanghai AI Laboratory \quad $^4$The Chinese University of Hong Kong \\
{\tt\small zx.zhang@sjtu.edu.cn} \\
}

\begin{document}
\maketitle
\begin{abstract}
Semi-supervised Video Object Segmentation aims to segment a specified target throughout a video sequence, initialized by a first-frame mask. Previous methods rely heavily on appearance-based pattern matching and thus exhibit limited robustness against challenges such as drastic visual changes, occlusions, and scene shifts. This failure is often attributed to a lack of high-level conceptual understanding of the target. The recently proposed \textbf{Se}gment \textbf{C}oncept (\textbf{SeC}) framework mitigated this limitation by using a Large Vision-Language Model (LVLM) to establish a deep semantic understanding of the object for more persistent segmentation. In this work, we evaluate its zero-shot performance on the challenging co\textbf{M}plex video \textbf{O}bject \textbf{SE}gmentation v2 (\textbf{MOSEv2}) dataset. Without any fine-tuning on the training set, SeC achieved 39.7 \JFn on the test set and ranked 2nd place in the Complex VOS track of the 7th Large-scale Video Object Segmentation Challenge.
\end{abstract}  

\section{Introduction}
Semi-supervised Video Object Segmentation (VOS) is a fundamental computer vision task that involves segmenting and tracking a specific object throughout a video, given the mask in the initial frame. This technique has wide-ranging applications in fields such as autonomous driving~\cite{siam2021video}, robotic perception~\cite{griffin2020video}, and video editing~\cite{tu2025videoanydoor}.

Previous VOS methods, such as Cutie~\cite{cheng2024putting} and SAM 2~\cite{ravi2025sam}, typically employ a memory-based paradigm,  
which constructs a memory bank from the visual features of previously segmented frames and subsequently queries this memory to guide the segmentation of new frames, achieving impressive performance on standard benchmarks~\cite{pont20172017, xu2018youtube, ding2023mose, hong2024lvos}. However, existing methods still show limitations when facing complex scenarios like severe occlusions, object disappearances, or abrupt shot changes, which may be due to their reliance on matching low-level appearance features and a lack of a deeper, conceptual understanding of the target object.

To mitigate this limitation, the recently proposed \textbf{Se}gment \textbf{C}oncept (\textbf{SeC})~\cite{zhang2025sec} model introduces a novel, concept-driven framework. SeC represents a paradigm shift from low-level pixel matching to high-level concept reasoning by using the capabilities of Large Vision-Language Models (LVLMs). The framework implicitly constructs a concept of the target by integrating visual cues across previous keyframes and integrating it into the segmentation pipeline. The SeC approach empowers the model to maintain robust tracking through semantic consistency, enabling it to handle complex scenarios such as severe occlusions, abrupt scene changes, and target disappearance-reappearance.

This report evaluates the zero-shot generalization capabilities of the SeC framework on the highly challenging MOSEv2 dataset~\cite{ding2025mosev2}. Without any fine-tuning on the training set, SeC achieved 39.7 \JFn on the test set, securing second place in the Complex VOS track of the 7th Large-scale Video Object Segmentation Challenge. In the following sections, we will detail our implementation, analyze the core technical advantages of the SeC model that contributed to this result, and discuss potential avenues for future research.

\begin{figure*}[t]
  \centering
  \includegraphics[width=0.9\textwidth]{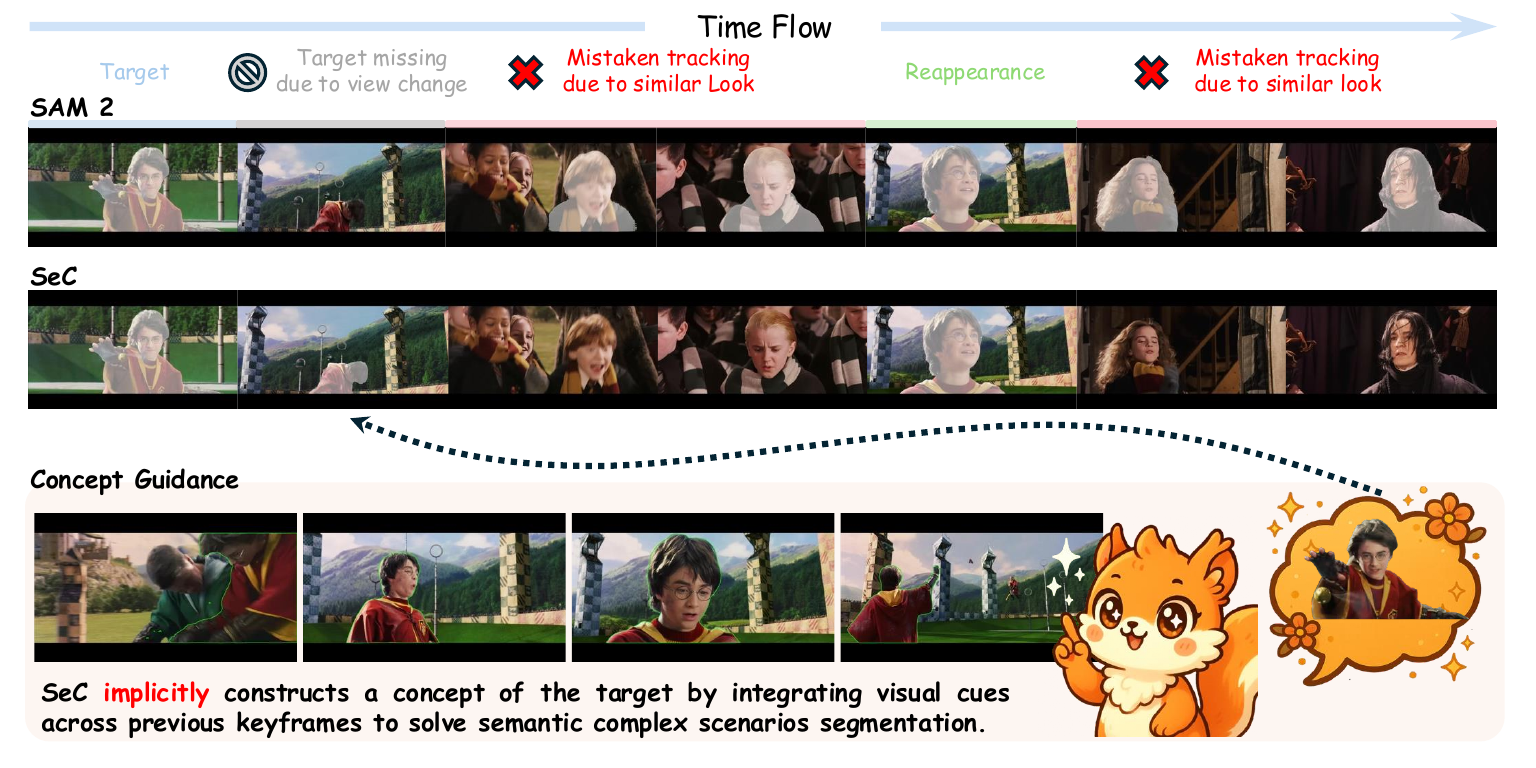}
  \vspace{-10pt}
  \caption{Overview of Segment Concept (SeC) framework. Image sourced from \cite{zhang2025sec}.}
  \label{fig:sec}
  \vspace{-10pt}
\end{figure*}

\section{Related Work}
\paragraph{Video Object Segmentation.}
Video Object Segmentation (VOS) focuses on the pixel-level segmentation of specific objects within a video, which is mainly evaluated under two primary settings~\cite{pont20172017, perazzi2016benchmark}: semi-supervised and unsupervised.
In the semi-supervised VOS setting, the model is provided with the target object's mask in the first frame and is required to track and segment that object throughout all subsequent frames, while unsupervised VOS automatically identifies and segments the most salient foreground objects by analyzing cues such as motion and appearance without any initial guidance.
A natural extension, Referring Video Object Segmentation (RVOS)~\cite{seo2020urvos}, replaces the initial mask prompt with a more flexible natural language description.
This tech report explores the application of SeC \cite{zhang2025sec} to semi-supervised VOS.

\paragraph{Semi-supervised Video Object Segmentation.} 
Previous Semi-supervised VOS methods typically employ a memory-based paradigm, relying on networks to match features between the query frame and historical frames. While classical models performed this matching at the pixel level~\cite{cheng2022xmem, yang2021associating}, more recent works have improved robustness by incorporating object-level information, leading to stronger target-background separation~\cite{wang2023look, cheng2024putting}.
A significant leap for the VOS field came with the advent of foundation models, particularly the Segment Anything Model 2 (SAM 2)~\cite{ravi2025sam}. By building a massive data engine that improves both model and data via user interaction, SAM 2 and its variants~\cite{yang2024samurai,ding2024sam2long, videnovic2025distractor} have achieved leading performance on the VOS task. Building on this progress, the SeC framework~\cite{zhang2025sec} introduced a concept reasoning paradigm designed to handle even more challenging scenarios.

\section{The LSVOS 2025 Challenge}
The 7th Large-scale Video Object Segmentation (LSVOS) Challenge, held in conjunction with ICCV 2025, aims to advance the state-of-the-art by benchmarking VOS models against complex and realistic scenarios. This year's challenge is structured into three distinct tracks, each designed to target a critical area of VOS research:
\begin{itemize}
    \item \textbf{Track 1: Complex Video Object Segmentation.} This track uses the challenging MOSEv2 dataset~\cite{ding2025mosev2} to assess model performance under a diverse range of extreme conditions, including adverse weather, low light, multi-shot sequences, camouflage, non-physical targets, and knowledge-dependent scenarios.
    \item \textbf{Track 2: Classic Video Object Segmentation.} This track benchmarks the traditional semi-supervised VOS task utilizing a combination of the LVOS~\cite{hong2023lvos,hong2024lvos} and MOSE~\cite{ding2023mose} datasets, which primarily focus on evaluating model performance in long-term scenarios and other complex scene dynamics.
    \item \textbf{Track 3: Referring Video Object Segmentation.} Using the MeViS dataset~\cite{ding2023mevis, ding2025mevis}, this track centers on language-guided segmentation. The key challenge is to identify target objects based on descriptions of their motion rather than their static visual attributes.
\end{itemize}

\section{Method}
We directly employed the Segment Concept (SeC) framework~\cite{zhang2025sec}. As shown in Figure~\ref{fig:sec}, SeC implicitly constructs a target concept from previous keyframes and integrates the conceptual reasoning capabilities of LVLMs with fine-grained pixel matching through a scene-adaptive activation strategy, which enables robust and efficient performance across complex scenarios.

\begin{figure*}[htbp]
  \centering
  \includegraphics[width=1.0\textwidth]{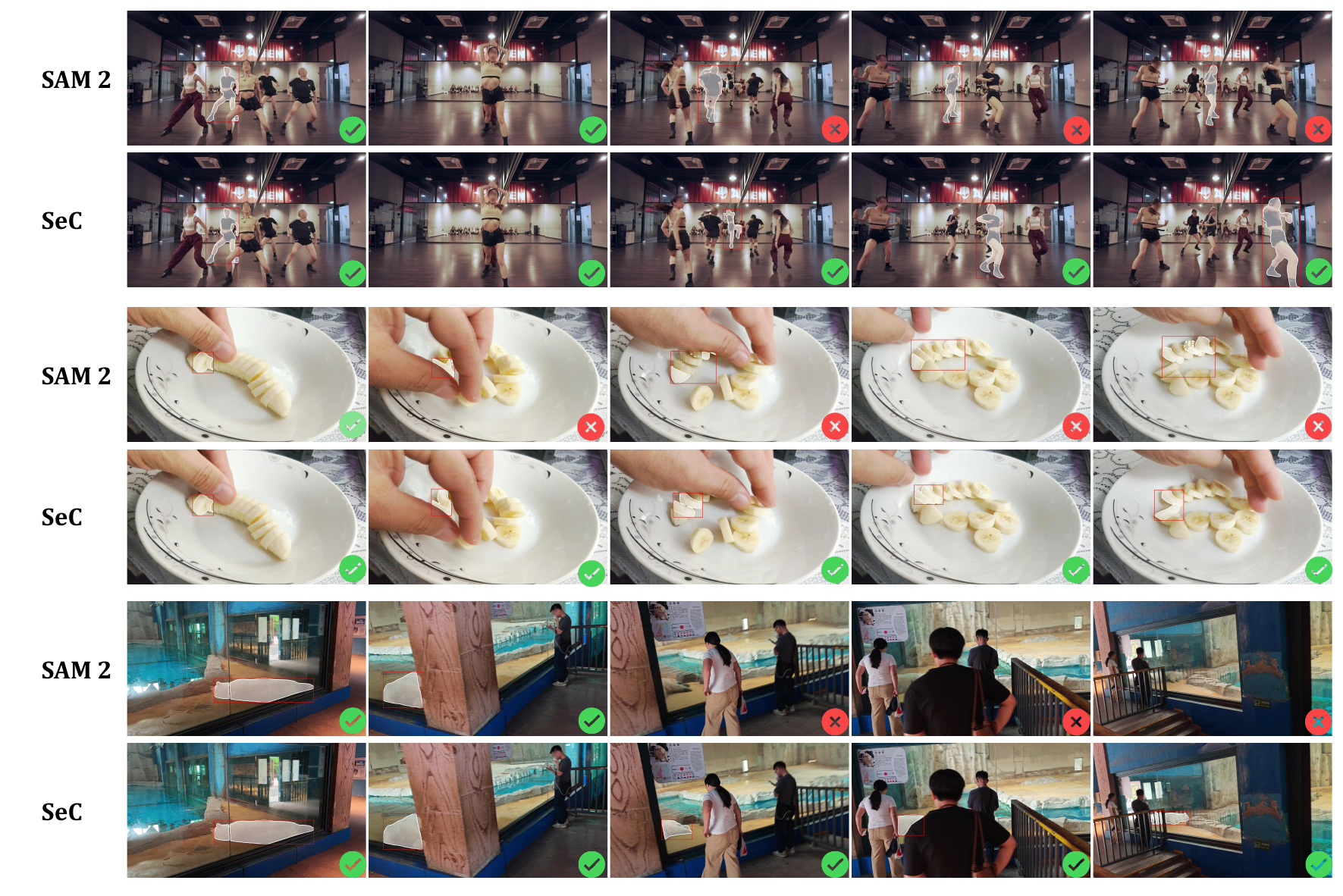}
  \caption{Qualitative comparison of SAM 2 and SeC on validation dataset. Image sourced from \cite{zhang2025sec}.}
  \label{fig:qualitative_examples}
\end{figure*}
\input{table_val}
\input{table_test}

\section{Experiment}
\subsection{Datasets and Metrics}
\paragraph{Dataset.} The dataset for this track is the challenging MOSEv2 dataset~\cite{ding2025mosev2}, which is designed to benchmark and advance VOS methods under realistic and complex conditions. MOSEv2 consists of 5,024 videos and over 701,976 high-quality masks for 10,074 objects across 200 categories. While amplifying the core challenges of its predecessor, such as frequent object disappearance-reappearance and severe occlusions, MOSEv2 introduces a range of new complexities. These include adverse weather (\textit{e.g.}, rain, snow), low-light scenes, multi-shot sequences, camouflaged objects, and non-physical targets like shadows and reflections. The dataset is officially partitioned into 3,666 training, 433 validation, and 614 testing videos. Our evaluation is primarily conducted on the official validation set and a designated 100-video subset of the test set.

\paragraph{Metrics.} Following \cite{ding2025mosev2}, we report the standard VOS metrics (\J, \F, and their average \JF) and additionally report \Fn, \JFn, \JFnd, and \JFnr. Among them, \JFn is selected as the primary evaluation metric.

\subsection{Implementation Details}
We directly test the released SeC-4B model~\cite{zhang2025sec} in a zero-shot setting, without any additional fine-tuning.

\subsection{Main Results}

As detailed in Table~\ref{tab:mosev2_val}, the SeC framework achieves leading performance on the MOSEv2 validation set, outperforming previous strong baselines, including several fine-tuned SAM2-L variants. In particular, SeC attains a primary score of 53.8 in \JFn, surpassing the SAM2Long-L by 2.3 points. Furthermore, Table~\ref{tab:mosev2_test} demonstrates that SeC achieved 39.7 \JFn on the test set and ranked 2nd place in the Complex VOS track of the 7th Large-scale Video Object Segmentation Challenge. Figure~\ref{fig:qualitative_examples} presents further qualitative visualization results, demonstrating SeC's robust segmentation performance in various complex scenarios.

\section{Conclusion}
In this report, we evaluate the zero-shot performance of the Segment Concept (SeC) framework on the challenging MOSEv2 dataset for the Complex VOS track of the 7th LSVOS Challenge. Without any fine-tuning on the training set, SeC achieved 39.7 \JFn on the test set and ranked 2nd place. The experimental results demonstrate the effectiveness of SeC's concept-guided paradigm, providing a reference for future VOS applications using SeC.

{
    \small
    \bibliographystyle{ieeenat_fullname}
    \bibliography{main}
}


\end{document}

%% file: table_val.tex
\begin{table*}[t]
\centering
\renewcommand\arraystretch{1.16}
\setlength{\tabcolsep}{4pt} % I slightly increased this for better spacing
% \footnotesize
\caption{The performance comparison of various baselines and SeC on the MOSEv2 validation set.
}
\label{tab:mosev2_val} % Changed the label to be more specific
\vspace{-2.16mm}
% The column specifier is now l|ccccccc for 1 label column + 7 data columns
\begin{tabular}{l|ccccccc}
\thickhline \\[-9pt]
\textbf{Method} & \JFn & \J & \Fn & \JFnd & \JFnr & \F & \JF \\[1pt]
\hline
\hline
SAM2-L~(ZS) & 49.5 & 47.7 & 51.3 & 62.9 & 27.3 & 53.6 & 50.7 \\
SAM2-L & 49.7 & 47.9 & 51.5 & 64.5 & 27.1 & 53.8 & 50.9 \\
SAMURAI-L & 51.1 & 49.0 & 53.2 & 52.4 & 34.9 & 55.8 & 52.4 \\
DAM4SAM-L & 51.2 & 49.2 & 53.2 & 57.2 & 34.2 & 55.6 & 52.4 \\
SAM2Long-L & 51.5 & 49.6 & 53.4 & 62.5 & 30.6 & 55.8 & 52.7 \\
\rowcolor[gray]{0.9}\textbf{SeC~(ZS)} & \textbf{53.8} & \textbf{51.9} & \textbf{55.7} & \textbf{70.4} & \textbf{34.1} & \textbf{58.4} & \textbf{55.2}\\
\thickhline
\end{tabular}
\vspace{-1mm}
\end{table*}

%% file: table_test.tex
\begin{table*}[t]
\centering
\renewcommand\arraystretch{1.16}
\setlength{\tabcolsep}{4pt}
\caption{The leaderboard of the MOSEv2 Track test set.}
\label{tab:mosev2_test}
\vspace{-2.16mm}
\begin{tabular}{l|ccccccc}
\thickhline \\[-9pt]
\textbf{Team} & \JFn & \J & \Fn & \JFnd & \JFnr & \F & \JF \\[1pt]
\hline
\hline
{mmm} & {39.89} & {39.02} & {40.76} & {57.15} & {19.0} & {42.35} & {40.68} \\
\rowcolor[gray]{0.9}qqqaaaaa & 39.7 & 38.87 & 40.53 & 57.84 & 18.67 & 42.09 & 40.48 \\
limjduni & 37.87 & 36.99 & 38.75 & 64.68 & 12.05 & 40.06 & 38.52 \\
waaaaaaaaa & 35.77 & 34.98 & 36.56 & 61.95 & 11.82 & 37.9 & 36.44 \\
springggg & 35.39 & 34.63 & 36.15 & 61.89 & 11.6 & 37.47 & 36.05 \\
\thickhline
\end{tabular}
\vspace{-1mm}
\end{table*}